\documentclass[11pt]{article}
\usepackage{silence}
\newif\ifanonymous
\anonymousfalse

\ifanonymous
  \usepackage[review]{acl}   
\else
  \usepackage{acl}           
\fi

\usepackage{microtype}
\usepackage{times}
\usepackage{latexsym}

\usepackage[T1]{fontenc}
\usepackage{inconsolata}

\usepackage{graphicx}
\usepackage{booktabs}
\usepackage{multirow}
\usepackage{amsmath,amssymb}
\usepackage{url}

\usepackage{tikz}
\usetikzlibrary{positioning,arrows.meta,shapes,fit}
\usepackage{pgfplots}
\pgfplotsset{compat=1.18}
\usepackage{tikz}
\usetikzlibrary{positioning,calc,arrows.meta}
\usepackage{placeins} 
\usepackage{algorithm}
\usepackage{algorithmic}
\usepackage{microtype}
\usepackage{booktabs}
\usepackage{makecell}

\title{Schema-Key Wording as an Instruction Channel in Structured Generation under Constrained Decoding}

\ifanonymous
\author{Anonymous ACL submission}
\else
\author{
  Yifan Le \\
  Independent Researcher \\
  \texttt{leyifan@zju.edu.cn}
}
\fi

\begin{document}
\maketitle

\begin{abstract}
Constrained decoding is widely used to make large language models produce
structured outputs that satisfy schemas such as JSON. Existing work mainly
treats schemas as structural constraints, overlooking that schema-key tokens
also enter the autoregressive context and may guide generation. To the best
of our knowledge, we present the first systematic study of schema keys as an
implicit instruction channel under constrained decoding. We formulate
structured generation as a multi-channel instruction problem, where task
signals can be placed in prompts, schema keys, or both. We further provide a projection-aware analysis that gives a sufficient
condition under which an unconstrained expected-score advantage of an
instructional key is preserved after grammar projection. Experiments on GSM8K and Math500 across seven
language models show that changing only schema-key wording can substantially
affect accuracy, with both positive and negative effects across models.
Prompt-level and schema-level instructions also interact non-additively.
The evidence is substantially stronger on GSM8K than on Math500. Our findings
show that schema design is not merely output formatting, but part of
instruction specification in structured generation.
\end{abstract}

\begin{figure*}[t]
    \centering
    \includegraphics[width=\textwidth]{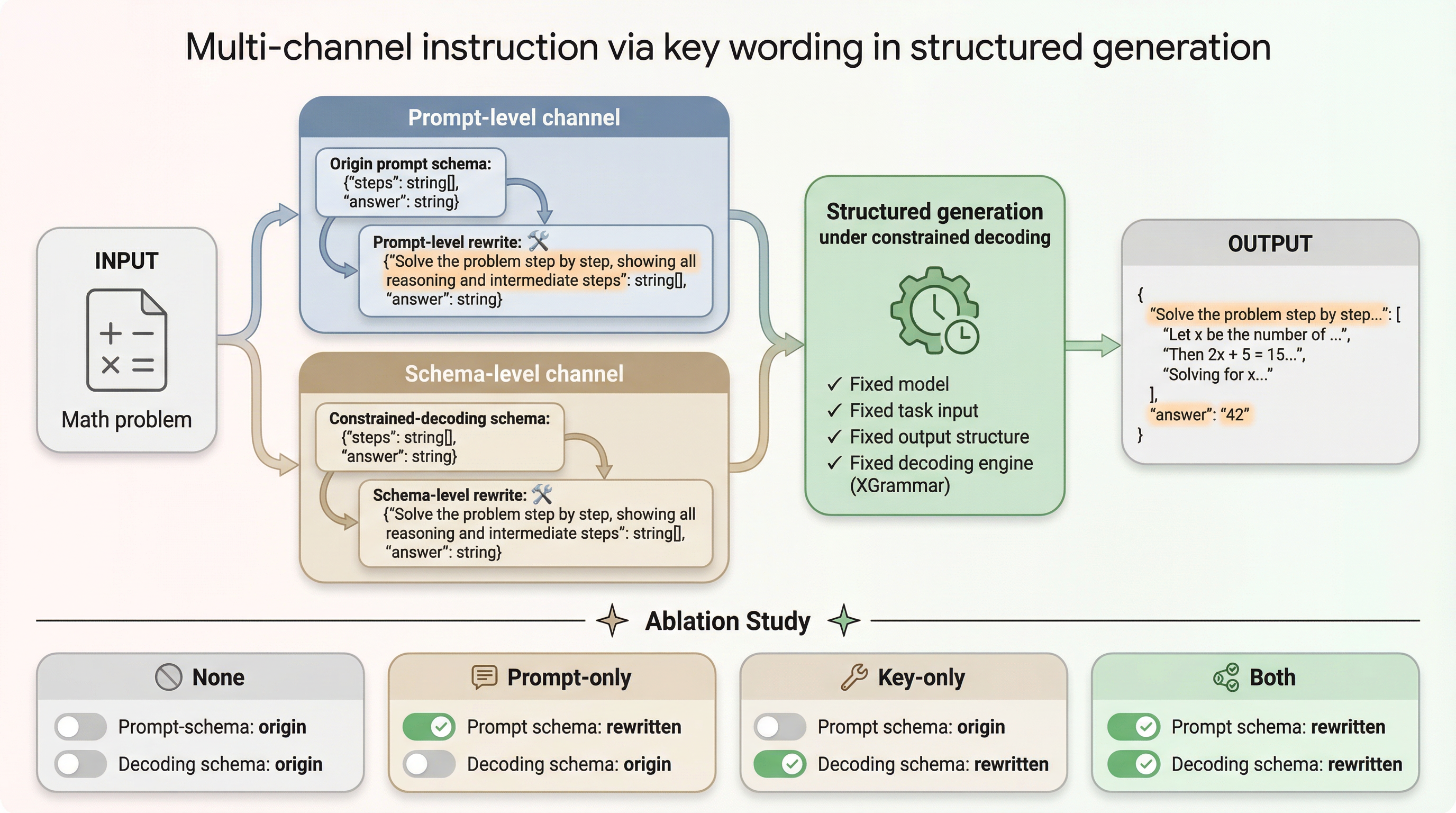}
    \caption{Overview of multi-channel instruction via key wording in structured generation. The same key wording can be injected through two locations: the schema string in the prompt and the schema used by constrained decoding. Holding the model, task input, output structure, and decoding engine fixed, we study four ablation settings: \textit{None}, \textit{Prompt-only}, \textit{Key-only}, and \textit{Both}.}
    \label{fig:overview}
\end{figure*}

\section{Introduction}

Structured generation is a key interface between large language models (LLMs) and downstream systems. In applications such as information extraction, tool use, and workflow automation, outputs often need to satisfy predefined formats such as JSON or schema-constrained records. Constrained decoding addresses this need by restricting next-token choices according to a grammar or schema, thereby ensuring structural validity \citep{scholak2021picard,willard2023efficient}.

However, structural validity is not the same as task success. Most constrained decoding work focuses on how constraints are represented and enforced, for example with context-free grammars, finite-state machines, or incremental parsers \citep{dong2024xgrammar,beurer2024guiding}. Recent studies further show that hard constraints can alter model behavior and sometimes affect reasoning quality or generation distribution \citep{ye2025efficient,yun2025price,wang2026draft,banerjee2025crane}. This suggests that structured generation should be studied not only as a validity problem, but also as an instruction-design problem.

We focus on a simple but overlooked source of instruction: schema keys. A JSON key is usually treated as a field name for parsing, but under autoregressive generation it is also a generated prefix that conditions the following value. Thus, a neutral key such as \texttt{steps} and a more instructional
key such as \texttt{Solve the problem step by step, showing all reasoning
and intermediate steps} can expose different semantic cues while preserving
the same canonical output structure. To the best of our knowledge, this is the first systematic study of schema keys as an implicit instruction channel under constrained decoding.

We formulate structured generation as a \textit{multi-channel instruction problem}, where instruction signals may be placed in the prompt, in schema keys, or in both. We compare four controlled settings: no additional instruction, key-only instruction, prompt-only instruction, and instruction through both channels. The model, task input, output structure, decoding engine, and hyperparameters are kept fixed, allowing us to isolate the effect of instruction placement.

To motivate why schema-key wording can matter, we provide a projection-aware analysis. Constrained decoding can be viewed as projecting the model's unconstrained distribution onto the grammar-valid distribution. Under this view, we derive a sufficient condition under which an
unconstrained expected-score advantage of an instructional key is preserved
after grammar projection. This provides a distribution-level perspective on
why schema-key effects may vary across models and settings.

Experiments on GSM8K and Math500 show that changing schema-key wording alone
can affect accuracy under constrained decoding. The direction and magnitude
of these effects vary substantially across individual models and tasks.
We also observe non-additive interactions between prompt-level and
schema-level instructions, indicating that the two channels may be
complementary, redundant, or conflicting depending on the model.
The evidence is particularly strong on GSM8K, whereas many Math500 effects
are smaller or statistically inconclusive. These results show that schema
design is not merely a formatting choice, but part of instruction
specification in structured generation.

Our contributions are as follows:
\begin{enumerate}
    \item We present the first systematic study of schema keys as an implicit instruction channel under constrained decoding.
    \item We introduce a multi-channel formulation of structured generation and isolate prompt-level, schema-level, and joint instruction effects through controlled ablations.
    \item We provide a projection-aware sufficient-condition analysis and
empirically show heterogeneous model- and task-dependent sensitivity to
schema-key and prompt-level instruction.
\end{enumerate}

\section{Related Work}

\subsection{Structured Generation with Constrained Decoding}

Structured generation is a key capability for real-world deployment of LLMs. Increasingly, benchmarks evaluate not only whether a model can follow instructions, but also whether it can produce outputs that satisfy explicit structural constraints \cite{zhou2023instruction,jiang2024followbench,liu2024we}. The dominant approach for structured output generation is constrained decoding, where candidate tokens are restricted during inference according to a predefined schema, ensuring that the final output is syntactically and structurally valid.

Most existing work in this area focuses on improving the efficiency and correctness of structural constraints. Representative methods rely on context-free grammars (CFGs), pushdown automata (PDAs), finite-state machines (FSMs), or incremental parsing to constrain model outputs \cite{dong2024xgrammar,mlc2026xgrammar2,willard2023efficient,park2025flexible,scholak2021picard,beurer2024guiding}. In addition, some work introduces extra generation stages or post-processing modules to preserve generation quality while maintaining structural validity, such as two-stage constrained decoding or dedicated models for structured transformation \cite{wang2025slot,wang2026draft,zheng2026thinking,banerjee2025crane}.

\subsection{Effects of Structural Constraints on Model Behavior}

Beyond structural validity and decoding efficiency, several studies have begun to examine how format constraints affect model behavior. Prior work shows that constrained decoding may introduce distributional bias and reduce diversity, creativity, or reasoning performance \cite{ye2025efficient,yun2025price,tam2024let,schall2025hidden}. At the same time, grammar constraints are not always harmful: in certain tasks, they can even improve model performance, although their effect depends heavily on model scale, task type, and constraint design \cite{raspanti2025grammar}. Overall, existing studies mainly analyze the side effects of constrained decoding from the perspectives of restricted search space, bias introduced by decoding constraints, or the strength of format control.

Unlike these studies, we focus on a finer-grained but largely overlooked factor: whether schema-key wording itself acts as an implicit instruction signal under constrained decoding.

\begin{table*}[t]
\centering
\small
\caption{
Performance under different instruction placements across models under greedy decoding.
``None'' corresponds to constrained decoding (CD) with neutral wording in both channels.
``Key only'' injects instruction through the decoder-enforced schema key,
``Prompt only'' injects instruction through the prompt-side schema,
and ``Both'' applies both channels.
All values are answer accuracy (\%).
The best result for each model and benchmark is shown in bold.
}
\label{tab:main_results}
\begin{tabular}{llcc}
\toprule
Model & Placement & GSM8K & Math500 \\
\midrule

\multirow{4}{*}{Qwen2.5-3B}
& None (CD)   & 75.06 & 34.20 \\
& Key only    & \textbf{79.38} & \textbf{37.00} \\
& Both        & 73.09 & 36.60 \\
& Prompt only & 77.63 & 35.00 \\
\midrule

\multirow{4}{*}{Qwen2.5-7B}
& None (CD)   & 81.27 & 39.40 \\
& Key only    & \textbf{89.08} & \textbf{42.20} \\
& Both        & 88.10 & 41.00 \\
& Prompt only & 84.15 & 41.20 \\
\midrule

\multirow{4}{*}{DeepSeek-R1-Distill-Qwen-7B}
& None (CD)   & 63.38 & 41.80 \\
& Key only    & 69.83 & 24.00 \\
& Both        & 76.72 & \textbf{47.40} \\
& Prompt only & \textbf{77.03} & 43.00 \\
\midrule

\multirow{4}{*}{Qwen2.5-Coder-14B}
& None (CD)   & 87.57 & \textbf{38.40} \\
& Key only    & 89.76 & 34.20 \\
& Both        & \textbf{91.13} & 34.60 \\
& Prompt only & 88.93 & 36.80 \\
\midrule

\multirow{4}{*}{Llama-3.2-1B}
& None (CD)   & 9.40 & 7.00 \\
& Key only    & 15.39 & 7.20 \\
& Both        & \textbf{23.73} & \textbf{12.60} \\
& Prompt only & 13.65 & 7.00 \\
\midrule

\multirow{4}{*}{Llama-3.2-3B}
& None (CD)   & 60.50 & 21.20 \\
& Key only    & 48.45 & 19.00 \\
& Both        & \textbf{68.46} & \textbf{22.60} \\
& Prompt only & 63.46 & 22.20 \\
\midrule

\multirow{4}{*}{Llama-3.1-8B}
& None (CD)   & 79.53 & 30.20 \\
& Key only    & 76.35 & \textbf{31.60} \\
& Both        & \textbf{82.79} & 31.20 \\
& Prompt only & 82.18 & 29.60 \\
\bottomrule
\end{tabular}
\end{table*}

\begin{figure}[t]
    \centering
    \includegraphics[width=\columnwidth]{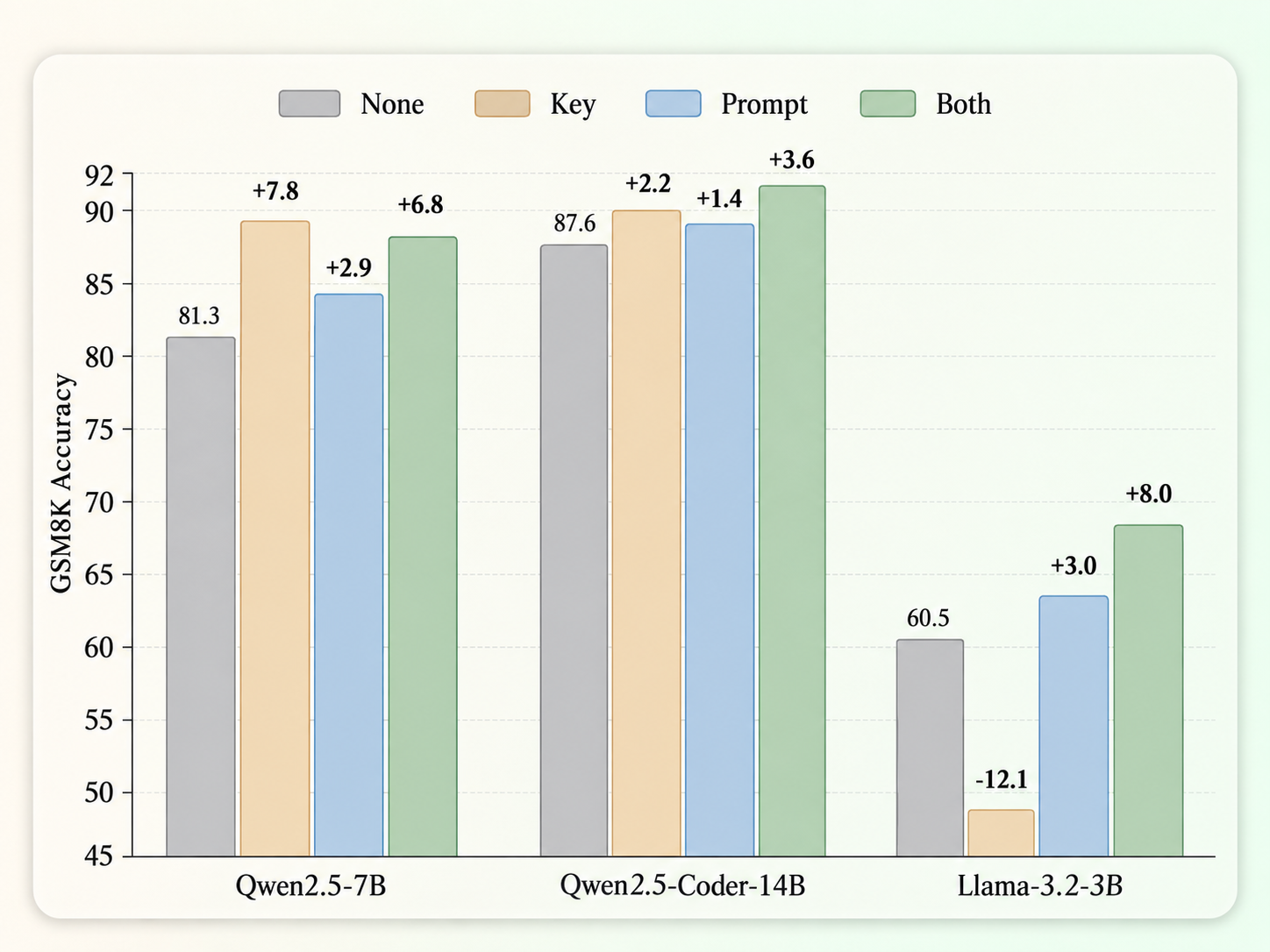}
    \caption{
Performance under different instruction placements on representative models in GSM8K.
Labels above the non-baseline bars indicate percentage-point changes relative to the \textit{None} setting.
The y-axis is truncated to improve visual comparability.
}
    \label{fig:rep_models}
\end{figure}

\begin{table*}[t]
\centering
\small
\caption{
Channel effects and prompt--schema interaction effects under greedy decoding.
Values are percentage-point changes with unadjusted 95\% paired item-level percentile bootstrap confidence intervals based on 20{,}000 resamples.
$\Delta_{\mathrm{key}}=R_{01}-R_{00}$,
$\Delta_{\mathrm{prompt}}=R_{10}-R_{00}$, and
$\Delta_{\mathrm{int}}=R_{11}-R_{10}-R_{01}+R_{00}$.
A superscript $^{*}$ indicates that the confidence interval excludes zero.
Effects are computed from item-level outcomes before rounding the accuracies shown in Table~\ref{tab:main_results}.
}
\label{tab:channel_effects}

\vspace{0.5em}

\textbf{(a) GSM8K}

\vspace{0.3em}

\begin{tabular}{lccc}
\toprule
Model
& $\Delta_{\mathrm{key}}$
& $\Delta_{\mathrm{prompt}}$
& $\Delta_{\mathrm{int}}$ \\
\midrule

Qwen2.5-3B
& $+4.32^{*}\,[+2.05,+6.60]$
& $+2.58^{*}\,[+0.38,+4.78]$
& $-8.87^{*}\,[-12.05,-5.76]$ \\

Qwen2.5-7B
& $+7.81^{*}\,[+5.84,+9.78]$
& $+2.88^{*}\,[+0.83,+4.93]$
& $-3.87^{*}\,[-6.37,-1.36]$ \\

DeepSeek-R1-Distill-Qwen-7B
& $+6.44^{*}\,[+3.41,+9.55]$
& $+13.65^{*}\,[+10.99,+16.38]$
& $-6.75^{*}\,[-10.61,-2.96]$ \\

Qwen2.5-Coder-14B
& $+2.20^{*}\,[+0.61,+3.79]$
& $+1.36\,[-0.23,+2.96]$
& $+0.00\,[-1.90,+1.97]$ \\

Llama-3.2-1B
& $+5.99^{*}\,[+3.64,+8.34]$
& $+4.25^{*}\,[+2.20,+6.29]$
& $+4.09^{*}\,[+0.83,+7.35]$ \\

Llama-3.2-3B
& $-12.05^{*}\,[-15.09,-9.02]$
& $+2.96^{*}\,[+0.30,+5.69]$
& $+17.06^{*}\,[+13.19,+21.00]$ \\

Llama-3.1-8B
& $-3.18^{*}\,[-5.38,-0.99]$
& $+2.65^{*}\,[+0.68,+4.62]$
& $+3.79^{*}\,[+1.06,+6.52]$ \\

\bottomrule
\end{tabular}

\vspace{0.8em}

\textbf{(b) Math500}

\vspace{0.3em}

\begin{tabular}{lccc}
\toprule
Model
& $\Delta_{\mathrm{key}}$
& $\Delta_{\mathrm{prompt}}$
& $\Delta_{\mathrm{int}}$ \\
\midrule

Qwen2.5-3B
& $+2.80\,[-0.80,+6.40]$
& $+0.80\,[-2.80,+4.40]$
& $-1.20\,[-6.20,+3.80]$ \\

Qwen2.5-7B
& $+2.80\,[-0.60,+6.20]$
& $+1.80\,[-1.40,+5.00]$
& $-3.00\,[-7.40,+1.40]$ \\

DeepSeek-R1-Distill-Qwen-7B
& $-17.80^{*}\,[-22.00,-13.60]$
& $+1.20\,[-2.40,+4.60]$
& $+22.20^{*}\,[+16.80,+27.80]$ \\

Qwen2.5-Coder-14B
& $-4.20^{*}\,[-7.40,-1.20]$
& $-1.60\,[-4.40,+1.20]$
& $+2.00\,[-1.80,+6.00]$ \\

Llama-3.2-1B
& $+0.20\,[-2.60,+3.00]$
& $+0.00\,[-2.60,+2.80]$
& $+5.40^{*}\,[+1.20,+9.60]$ \\

Llama-3.2-3B
& $-2.20\,[-5.80,+1.40]$
& $+1.00\,[-2.20,+4.40]$
& $+2.60\,[-2.20,+7.40]$ \\

Llama-3.1-8B
& $+1.40\,[-1.80,+4.60]$
& $-0.60\,[-3.80,+2.60]$
& $+0.20\,[-4.40,+5.00]$ \\

\bottomrule
\end{tabular}

\end{table*}

\section{Method}
\label{sec:method}

\subsection{Problem Setup}

We study structured generation where a model receives an input $x$ and generates a serialized output $y$ that must satisfy a schema $s$, such as a JSON schema. Constrained decoding enforces this requirement by masking invalid next tokens. At step $t$, let the unconstrained distribution be
\begin{equation}
    p_t(v)=p_\theta(v\mid x,y_{<t}).
\end{equation}
Given the valid token set $\mathcal{V}_t(s,y_{<t})$, constrained decoding uses
\begin{equation}
\begin{aligned}
    q_t(v)
    &=
    \frac{p_t(v)\mathbf{1}[v\in\mathcal{V}_t(s,y_{<t})]}
    {Z_t(s,y_{<t})}, \\
    Z_t(s,y_{<t})
    &=
    \sum_{u\in\mathcal{V}_t(s,y_{<t})}p_t(u).
\end{aligned}
\label{eq:cd_distribution}
\end{equation}
Although this guarantees syntactic validity, it does not make the schema semantically neutral. In autoregressive generation, schema keys appear before their values and thus condition subsequent value generation.

Our intervention changes only the wording of a target key. Let $s_k$ be a schema whose target field is named by key $k$, and let $\pi_k$ map valid outputs under $s_k$ to the same canonical object used for evaluation. Thus, different key wordings can preserve the same field order, value types, number of fields, and canonical interpretation while changing the generated key tokens.

\subsection{Projection-Aware View of Schema-Key Instructions}
\label{sec:projection_view}

Following the projection view of constrained decoding \citep{wang2026draft}, we characterize when an instructional key can help. Let $k_0$ denote the neutral key \texttt{steps}, and let $k_1$ denote
the instructional key used in our experiments:
\texttt{Solve the problem step by step, showing all reasoning and
intermediate steps}. After the key prefix has appeared in the context, let $\bar{P}_{k,x}$ denote the unconstrained continuation distribution over the value field, and let $\bar{Q}_{k,x}$ denote the corresponding distribution after grammar projection. This conditional view factors out the mechanical cost of forcing the key string itself and focuses on its downstream effect on value generation.

For a bounded metric $M\in[0,1]$, define
\begin{equation}
\tilde{M}_k(y)=
\begin{cases}
M(\pi_k(y)), & y\in\mathcal{L}(s_k),\\
0, & y\notin\mathcal{L}(s_k),
\end{cases}
\label{eq:bounded_metric}
\end{equation}
and define the expected scores
\begin{equation}
\begin{aligned}
    R_{\bar{P}}(k,x)
    &=\mathbb{E}_{y\sim\bar{P}_{k,x}}[\tilde{M}_k(y)], \\
    R_{\bar{Q}}(k,x)
    &=\mathbb{E}_{y\sim\bar{Q}_{k,x}}[\tilde{M}_k(y)].
\end{aligned}
\label{eq:expected_scores}
\end{equation}
Let the value-level projection tax and its induced bound be
\begin{equation}
\begin{aligned}
    \mathcal{T}_{\mathrm{val}}(k,x)
    &=D_{\mathrm{KL}}(\bar{Q}_{k,x}\|\bar{P}_{k,x}), \\
    B_k(x)
    &=\sqrt{\tfrac{1}{2}\mathcal{T}_{\mathrm{val}}(k,x)}.
\end{aligned}
\label{eq:tax_bound_acl}
\end{equation}
Under the standard shared-support and stopping convention, Pinsker's inequality gives
\begin{equation}
    \left|R_{\bar{Q}}(k,x)-R_{\bar{P}}(k,x)\right|
    \le B_k(x).
\label{eq:pinsker_method}
\end{equation}
Therefore, if
\begin{equation}
\begin{aligned}
R_{\bar{P}}(k_1,x)-R_{\bar{P}}(k_0,x)
> B_{k_1}(x)+B_{k_0}(x),
\end{aligned}
\label{eq:sufficient_condition_key}
\end{equation}
then $R_{\bar{Q}}(k_1,x)>R_{\bar{Q}}(k_0,x)$. This gives a sufficient condition under which an unconstrained
expected-score advantage of an instructional key is preserved after grammar
projection. The condition is sufficient but not necessary. The full
derivation is in Appendix~\ref{app:projection_derivation}.

This analysis is distributional. Since our experiments use greedy decoding, the bound should be interpreted as a diagnostic explanation rather than a deterministic guarantee for every generated trajectory.

\subsection{Structured Generation as Multi-Channel Instruction}

We formulate structured generation as a multi-channel instruction problem. Let $c_p$ denote instruction signals placed in the prompt, and let $c_s$ denote instruction signals placed in schema keys enforced by the decoder:
\begin{equation}
    y \sim p_\theta(\cdot\mid x,c_p,c_s;s).
\end{equation}
Here, $s$ specifies the structural constraint, while $c_p$ and $c_s$ specify where semantic guidance is injected.

We compare four instruction-placement settings:
\begin{itemize}
    \item \textbf{None}: both the prompt schema and decoder-enforced schema use neutral keys.
    \item \textbf{Key-only}: the prompt schema uses neutral keys, while the decoder-enforced schema uses an instructional key.
    \item \textbf{Prompt-only}: the prompt schema describes an instructional key, while the decoder-enforced schema keeps the neutral key.
    \item \textbf{Both}: the instructional key is used in both the prompt schema and the decoder-enforced schema.
\end{itemize}
This separates whether the instruction is only described to the model, forced into the generated output, or supplied through both routes.

\subsection{Controlled Comparison and Effect Measures}

Across settings, we keep the model, dataset, constrained decoding engine, decoding hyperparameters, model parameters, field order, value types, number of fields, and canonical interpretation fixed. The only intended intervention is the placement of the instruction signal.

Let $R_{ij}$ denote the evaluation result when $c_p=i$ and $c_s=j$, where $i,j\in\{0,1\}$. We quantify the schema-level, prompt-level, and joint effects as
\begin{equation}
\begin{aligned}
\Delta_{\mathrm{key}} &= R_{01}-R_{00}, \\
\Delta_{\mathrm{prompt}} &= R_{10}-R_{00}, \\
\Delta_{\mathrm{both}} &= R_{11}-R_{00}.
\end{aligned}
\label{eq:channel_effects}
\end{equation}
To measure whether the two channels combine additively, we define
\begin{equation}
    \Delta_{\mathrm{int}}
    =R_{11}-R_{10}-R_{01}+R_{00}.
\label{eq:interaction_effect}
\end{equation}
Positive values indicate synergy, near-zero values indicate approximate additivity, and negative values indicate redundancy or conflict between prompt-level and schema-level instructions.

\section{Experiments}
\label{sec:experiments}

\subsection{Experimental Setup}

\paragraph{Tasks and metrics.}
We evaluate instruction placement on two mathematical reasoning benchmarks, \textbf{GSM8K}~\cite{cobbe2021training} and \textbf{Math500}~\cite{hendrycks2021measuring}. Both tasks require multi-step reasoning and a final answer, making them suitable for studying whether schema keys influence reasoning behavior under constrained decoding. We parse the generated JSON output, extract the final answer field, and report answer accuracy.

As a decoding-regime sensitivity analysis, we additionally repeat the
original stochastic decoding configuration with \texttt{temperature}=0.8
for three complete runs while keeping all other inference settings fixed.
The results are reported in
Appendix~\ref{app:temperature_sensitivity}.

\paragraph{Models.} To study cross-model sensitivities to instruction channels, we evaluate a diverse set of representative open-source models, covering both the Qwen family \cite{qwen2024qwen25,qwen2024qwen25coder} and the LLaMA family \cite{grattafiori2024llama3,meta2024llama32}, as well as a reasoning-oriented variant based on Qwen, DeepSeek-R1-Distill-Qwen-7B \cite{deepseek2025r1distill}. Specifically, we evaluate Qwen2.5-3B, Qwen2.5-7B, DeepSeek-R1-Distill-Qwen-7B, Qwen2.5-Coder-14B, Llama3.2-1B, Llama-3.2-3B, and Llama-3.1-8B. This model set covers different parameter scales, model families, and post-training styles, allowing us to examine whether schema-key effects are stable across models or tied to particular instruction-following behaviors.

\begin{table}[t]
\centering
\small
\caption{
Effect of instruction placement across channels on representative models on GSM8K under greedy decoding.
For clarity, we report a subset of models illustrating heterogeneous responses to instruction placement.
These examples are intended to illustrate model-level variation rather than a family-level trend.
The best result for each model is shown in bold.
}
\label{tab:subset_results}
\begin{tabular}{lcccc}
\toprule
Model & None & Key & Prompt & Both \\
\midrule
Qwen2.5-7B
& 81.27
& \textbf{89.08}
& 84.15
& 88.10 \\

Qwen2.5-Coder-14B
& 87.57
& 89.76
& 88.93
& \textbf{91.13} \\

Llama-3.2-3B
& 60.50
& 48.45
& 63.46
& \textbf{68.46} \\
\bottomrule
\end{tabular}
\end{table}

\paragraph{Structured output and constrained decoding.}
All experiments use the same structured generation framework and the same constrained decoding backend, \textbf{XGrammar}~\cite{dong2024xgrammar}. The target JSON output contains a reasoning field and an answer field. Across all settings, we keep the number of fields, field order, value types, parsing procedure, answer extraction rule, decoding backend, and inference configuration fixed. The controlled intervention is where the reasoning instruction is placed: in the prompt-side schema description, in the decoder-enforced schema key, or in both.

All main experiments use greedy decoding with vLLM and XGrammar.
We use \texttt{temperature}=0, \texttt{n}=1,
\texttt{top\_p}=1.0, \texttt{top\_k}=0,
\texttt{min\_p}=0.0,
\texttt{presence\_penalty}=0,
\texttt{frequency\_penalty}=0,
\texttt{repetition\_penalty}=1.0,
\texttt{max\_tokens}=8192, and \texttt{seed}=None.
For statistical uncertainty, we report unadjusted 95\% paired item-level
percentile bootstrap confidence intervals based on 20{,}000 resamples.

\paragraph{Instruction-placement settings.}
Following Section~\ref{sec:method}, we compare four settings: \textbf{None}, \textbf{Key-only}, \textbf{Prompt-only}, and \textbf{Both}. In None, both channels use neutral schema wording. In Key-only, the instructional wording appears only in the decoder-enforced schema key. In Prompt-only, the same instruction is described in the prompt while the enforced key remains neutral. In Both, the instruction appears in both locations. This design separates whether the instruction is merely described to the model, forced into the generated context, or supplied through both channels.

\subsection{Main Results under Different Instruction Placements}

Table~\ref{tab:main_results} reports the full results across models and benchmarks. The central observation is that schema-key wording is not a neutral formatting detail under constrained decoding. Even when the output structure and decoding engine are fixed, changing the decoder-enforced key can noticeably change task accuracy. This supports our view that schema keys can become part of the effective instruction context once they are generated autoregressively.

On GSM8K, Key-only improves Qwen2.5-7B from 81.27 to 89.08 and
Qwen2.5-3B from 75.06 to 79.38. The effect is not uniformly positive:
Llama-3.2-3B drops from 60.50 to 48.45 under Key-only, whereas
Llama-3.2-1B improves from 9.40 to 15.39.
DeepSeek-R1-Distill-Qwen-7B also improves under Key-only from 63.38 to
69.83 and reaches 77.03 under Prompt-only.
These results show that schema-level instruction is a real behavioral
factor, but its direction and magnitude depend on the model and task.

The Both setting further shows that prompt-level and schema-level
instructions are not uniformly additive. On GSM8K, Both yields the best
performance for Qwen2.5-Coder-14B, Llama-3.2-1B, Llama-3.2-3B, and
Llama-3.1-8B, whereas Qwen2.5-7B performs best under Key-only and
DeepSeek-R1-Distill-Qwen-7B performs best under Prompt-only.
Thus, combining the same instruction across both channels can lead to
complementarity, redundancy, or interference depending on the model.

Overall, Table~\ref{tab:main_results} supports three claims: schema keys can act as instruction-bearing tokens; instruction-channel sensitivity is model- and task-dependent; and prompt-level and schema-level signals can interact rather than simply reinforce each other.

\subsection{Cross-Model Sensitivity to Instruction Channels}

Table~\ref{tab:subset_results} highlights representative GSM8K cases that make the cross-model pattern easier to inspect. Unlike Table~\ref{tab:main_results}, which provides the complete benchmark results, this subset isolates three distinct behaviors: Qwen2.5-7B benefits most from schema-level instruction, Qwen2.5-Coder-14B obtains its best result when both channels are used, and Llama-3.2-3B is harmed by Key-only but improves under prompt-side guidance.

These cases suggest that the effectiveness of instruction placement is not determined only by the benchmark or by the formal JSON schema. It also depends on how a model interprets text appearing in different parts of the structured generation interface. 
A schema key that helps one model can distract another, even when the
constrained output space is identical.
Figure~\ref{fig:rep_models} provides a visual summary of these representative patterns.

\subsection{Single-Channel and Interaction Effects}

\begin{figure*}[t]
    \centering
    \includegraphics[width=0.96\textwidth]{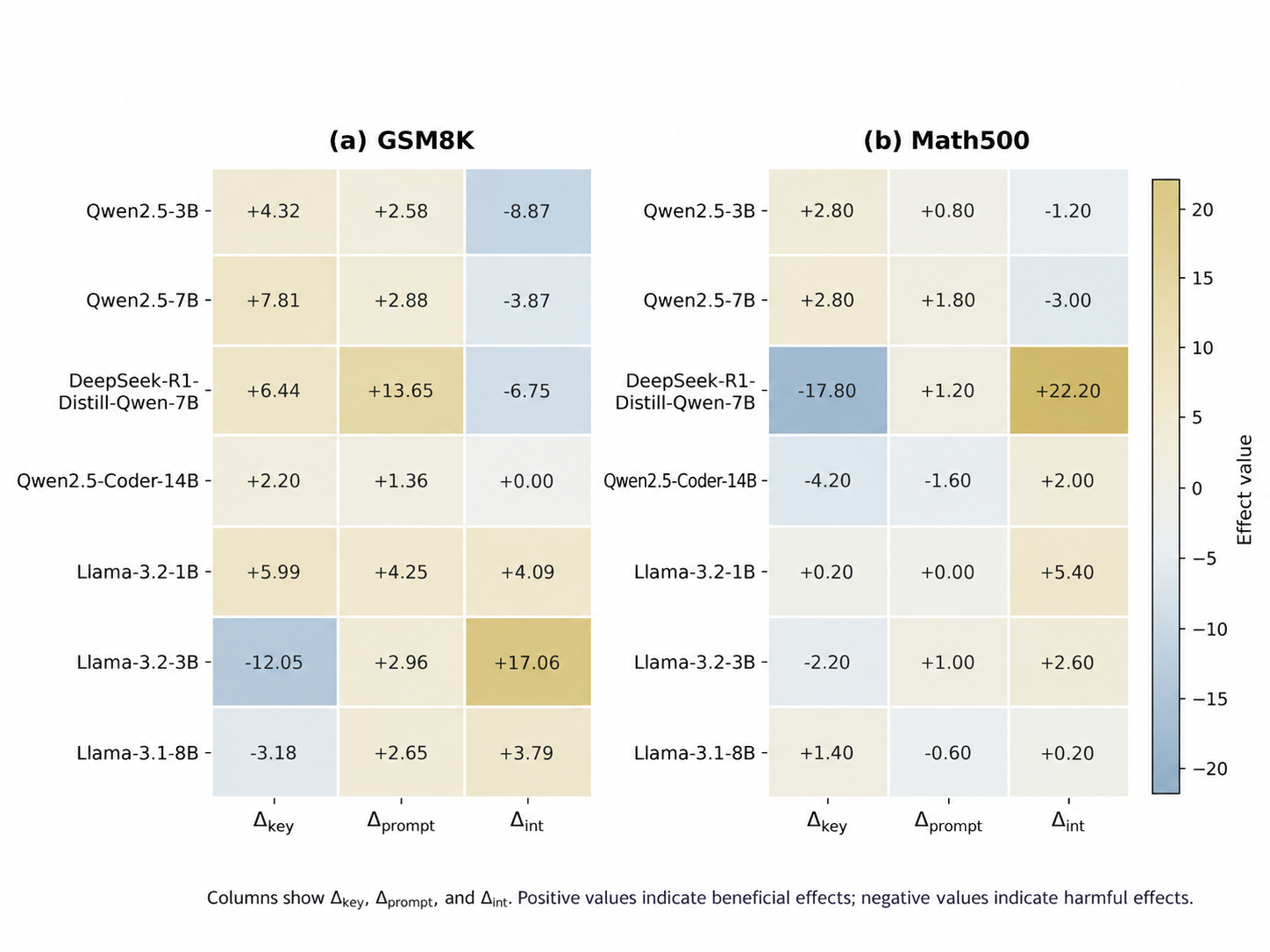}
    \caption{
    Channel-sensitivity map of single-channel and interaction effects on GSM8K and Math500.
    Rows correspond to models, and columns correspond to $\Delta_{\mathrm{key}}$, $\Delta_{\mathrm{prompt}}$, and $\Delta_{\mathrm{int}}$.
    Positive values indicate beneficial effects, while negative values indicate harmful effects.
    The map highlights model- and task-dependent sensitivity to instruction
placement and the non-additive interaction between prompt-level and
schema-level instructions.
    }
    \label{fig:channel_map}
\end{figure*}

To connect the empirical results with the effect definitions in Section~3, Table~\ref{tab:channel_effects} reports both single-channel effects and interaction effects on GSM8K and Math500. Figure~\ref{fig:channel_map} visualizes the same quantities as a channel-sensitivity map, making the cross-model and cross-benchmark patterns easier to compare.

As shown in Figure~\ref{fig:channel_map}, instruction-placement effects are
heterogeneous across models and tasks. On GSM8K, the paired bootstrap
confidence intervals exclude zero for schema-key effects in all seven
models, with five positive and two negative effects. Prompt effects are
positive and supported for six of seven models, while interaction effects
are supported for six of seven models and can be either positive or
negative. For example, Qwen2.5-7B shows a positive key effect
($\Delta_{\mathrm{key}}=+7.81$) and a negative interaction
($\Delta_{\mathrm{int}}=-3.87$), whereas Llama-3.2-3B shows a negative
key effect ($-12.05$) and a large positive interaction ($+17.06$).
DeepSeek-R1-Distill-Qwen-7B shows positive key and prompt effects but a
negative interaction. In contrast, Math500 provides weaker evidence:
all prompt main effects are statistically inconclusive, and only selected
key and interaction effects have confidence intervals excluding zero.

For Qwen2.5-7B, all instructed settings improve over None, with the largest gain coming from Key-only. Qwen2.5-Coder-14B also improves under all instructed settings, but obtains its largest gain under Both, indicating a more complementary relationship between prompt-level and schema-level instruction. In contrast, Llama-3.2-3B suffers a large negative change under Key-only but improves under Prompt-only and Both. Llama-3.1-8B is harmed by Key-only while improving under Prompt-only and Both. These results explain why schema wording should be treated as a model-dependent instruction design choice rather than a purely syntactic field name.

The same table also helps interpret the interaction between channels.
If prompt-level and schema-level instructions combined additively,
$\Delta_{\mathrm{int}}$ would be approximately zero.
Instead, Table~\ref{tab:channel_effects} shows both positive and negative
supported interaction effects on GSM8K, indicating that the two channels
can exhibit synergy, redundancy, or interference depending on the model.
For Qwen2.5-7B, the interaction is negative, whereas
Qwen2.5-Coder-14B is approximately additive and Llama-3.2-3B shows a
strong positive interaction.
These patterns support the multi-channel view introduced in Section~\ref{sec:method}: constrained decoding does not merely enforce output validity, but also changes where instruction signals enter the generation process. Prompt descriptions and schema keys can reinforce each other, compete with each other, or produce asymmetric effects depending on the model.

\subsection{Summary of Findings}

The experiments lead to three main findings.
First, schema-key wording can substantially influence reasoning accuracy
even when the output structure and constrained decoding engine are fixed.
Second, the direction and magnitude of these effects vary across individual
models and tasks, with substantially stronger statistical evidence on
GSM8K than on Math500.
Third, prompt-level and schema-level instructions can interact
non-additively, with both positive and negative interaction effects observed
across models.
These findings suggest that schema keys should be treated as
instruction-bearing design choices rather than neutral field names in
structured generation.

\section{Discussion and Conclusion}
\label{sec:discussion_conclusion}

This work shows that structured generation under constrained decoding is not only a matter of enforcing valid output formats. Because schema keys are generated as part of the autoregressive context, their wording can act as an instruction-bearing signal rather than a neutral parsing detail. Our results demonstrate that changing the reasoning-field key can alter downstream accuracy even when the schema structure, field types, parsing rules, and decoding backend are fixed.

The multi-channel formulation helps interpret these model-dependent and
non-additive effects. Prompt-side descriptions and decoder-enforced keys inject instruction signals through different routes: the prompt provides guidance before generation, while the schema key is forced into the generated prefix and directly conditions the following value. As a result, placing the same instruction in the prompt, in the schema key, or in both can lead to different outcomes. The experiments show substantial heterogeneity across individual models:
schema-level instruction can help, hurt, or interact with prompt-side
guidance depending on the model and task. The interaction analysis further shows that the two channels can be complementary, redundant, or even interfering, rather than simply additive.

These findings suggest a practical implication for structured generation systems: schema wording should be treated as part of the instruction design. Developers often tune prompts while leaving field names as arbitrary implementation choices, but our results show that this separation is incomplete under constrained decoding. For fair evaluation and reproducible deployment, prompts, schemas, key names, decoding backends, and parsing rules should be reported together. In conclusion, schema keys do not merely describe the shape of an answer; under constrained decoding, they can also shape how the answer is produced.

\section*{Limitations}

Our study focuses on mathematical reasoning benchmarks, specifically GSM8K and Math500. These tasks are useful for studying intermediate reasoning and final-answer extraction, but they do not cover the full range of structured generation scenarios, such as information extraction, tool use, code generation, data-to-text generation, or agent workflows. Our model coverage is also limited to several open-source Qwen and Llama-family models, together with a Qwen-based distilled reasoning model, so the observed instruction-channel sensitivities may not directly transfer to other model families, larger systems, or proprietary models. Finally, we study a controlled but limited schema design space centered on reasoning-field wording. Other schema factors, including field descriptions, field order, nesting depth, optional fields, enum values, and natural-language schema documentation, may also affect model behavior and deserve further study.

Our evaluated JSON schema contains two required fields and remains
structurally simple; we do not test nested schemas, optional fields, or a
broader range of key paraphrases. In addition, we do not include an
unconstrained-generation baseline or directly measure the terms in the
projection-aware analysis. Finally, the statistical evidence is
substantially stronger on GSM8K than on Math500, where many effects remain
inconclusive.
\section*{Ethics Statement}

This work uses publicly available benchmarks and open-source models, and does not involve private user data, human subjects, or personally identifiable information. The main ethical consideration is transparency in structured-generation evaluation. Since schema key wording can influence model performance, benchmark results under constrained decoding may depend on implementation choices that are easy to overlook. We therefore recommend that structured-generation studies report not only prompts and models, but also schemas, key names, decoding backends, and parsing rules. In deployment, schema variants should be evaluated carefully, especially in high-stakes settings where subtle changes in field names may affect correctness, robustness, or user trust.

\bibliography{custom}

\appendix

\section{Derivation of the Projection-Aware Schema-Key View}
\label{app:projection_derivation}

This appendix provides the derivation behind the projection-aware view in Section~\ref{sec:projection_view}. The analysis is distributional. Our main experiments use greedy decoding, so the result should be interpreted as a theoretical diagnostic of constraint-induced distortion rather than a deterministic guarantee for every greedy trajectory.

\subsection{Constrained Decoding as KL Projection}

At step $t$, let $p_t$ be the unconstrained next-token distribution and let $\mathcal{V}_t$ be the valid next-token set under the grammar and current prefix. The constrained distribution is
\begin{equation}
\begin{aligned}
    q_t(v)
    &=
    \frac{p_t(v)\mathbf{1}[v\in\mathcal{V}_t]}{Z_t}, \\
    Z_t
    &=
    \sum_{v\in\mathcal{V}_t}p_t(v).
\end{aligned}
\end{equation}
For any distribution $r$ supported on $\mathcal{V}_t$, we have
\begin{equation}
\begin{aligned}
D_{\mathrm{KL}}(r\|p_t)
&=
\sum_{v\in\mathcal{V}_t}
 r(v)\log\frac{r(v)}{p_t(v)} \\
&=
\sum_{v\in\mathcal{V}_t}
 r(v)\log\frac{r(v)}{Z_tq_t(v)} \\
&=
D_{\mathrm{KL}}(r\|q_t)-\log Z_t.
\end{aligned}
\label{eq:kl_projection_acl}
\end{equation}
Since $D_{\mathrm{KL}}(r\|q_t)\ge 0$, the minimum is achieved at $r=q_t$, with value $-\log Z_t$. Thus, constrained decoding performs a reverse-KL projection of the model distribution onto the set of valid next-token distributions.

The token-level projection cost is therefore
\begin{equation}
D_{\mathrm{KL}}(q_t\|p_t)=-\log Z_t.
\end{equation}
For a sequence $y=(y_1,\ldots,y_T)$ sampled under the constrained distribution $Q$, assuming the same stopping convention for $P$ and $Q$ and $Q(y)>0\Rightarrow P(y)>0$, the sequence-level KL is
\begin{equation}
\begin{aligned}
D_{\mathrm{KL}}(Q\|P)
&=
\mathbb{E}_{y\sim Q}
\left[
\log \frac{Q(y\mid x)}{P(y\mid x)}
\right] \\
&=
\mathbb{E}_{y\sim Q}
\left[
\sum_{t=1}^{T}
\log
\frac{q_t(y_t)}{p_t(y_t)}
\right] \\
&=
\mathbb{E}_{y\sim Q}
\left[
\sum_{t=1}^{T}
-\log Z_t(G_k,h_t)
\right].
\end{aligned}
\label{eq:seq_kl_acl}
\end{equation}
We call this quantity the cumulative projection tax:
\begin{equation}
\mathcal{T}(G_k,x)
=
\mathbb{E}_{y\sim Q}
\left[
\sum_{t=1}^{T}
-\log Z_t(G_k,h_t)
\right].
\end{equation}
The projection tax is small when the model already assigns high probability mass to valid continuations, and large when constrained decoding must strongly override the model's preferred next-token distribution.

\subsection{Schema-Key Conditioning}

Let $s(k)$ denote the token sequence corresponding to schema key $k$. In JSON-style constrained decoding, the key tokens are generated before the corresponding value. Hence, after the field name has been emitted, the model distribution over the value is conditioned on $s(k)$:
\begin{equation}
    p_\theta(y_{\mathrm{val}}\mid x,h,s(k)).
\end{equation}
This provides a direct autoregressive conditioning pathway through which
schema-key wording can affect generation. An instructional key may shift
the continuation distribution relative to a neutral key. A CoT-style key can move the model distribution toward reasoning-like continuations, while a neutral key may provide weaker semantic guidance.

For this reason, we define a conditional reference distribution $\bar{P}_{k,x}$ over continuations after the key prefix $s(k)$ is already present in the context. We also define $\bar{Q}_{k,x}$ as the corresponding constrained continuation distribution after the same prefix. This conditional formulation isolates the semantic effect of the key from the structural cost of forcing the key tokens themselves.

\subsection{Bounding the Effect of Projection}

Let $\tilde{M}_k(y)\in[0,1]$ be the bounded metric defined in Section~\ref{sec:projection_view}. For any two distributions $A$ and $B$ over the same continuation space,
\begin{equation}
\begin{aligned}
&\left|
\mathbb{E}_{y\sim A}[\tilde{M}_k(y)]
-
\mathbb{E}_{y\sim B}[\tilde{M}_k(y)]
\right| \\
&\hspace{3em}\le
\mathrm{TV}(A,B),
\end{aligned}
\end{equation}
where $\mathrm{TV}$ denotes total variation distance. By Pinsker's inequality,
\begin{equation}
\mathrm{TV}(A,B)
\le
\sqrt{\tfrac{1}{2}D_{\mathrm{KL}}(A\|B)}.
\end{equation}
Taking $A=\bar{Q}_{k,x}$ and $B=\bar{P}_{k,x}$ gives
\begin{equation}
\left|
R_{\bar{Q}}(k,x)-R_{\bar{P}}(k,x)
\right|
\le
B_k(x),
\end{equation}
with $B_k(x)$ defined in Equation~\ref{eq:tax_bound_acl}.

\subsection{A Sufficient Condition for CoT-Key Improvement}

Let $k_0$ be a neutral key and $k_1$ be a CoT-style key. From the bound above,
\begin{equation}
\begin{aligned}
R_{\bar{Q}}(k_1,x)
&\ge
R_{\bar{P}}(k_1,x)-B_{k_1}(x), \\
R_{\bar{Q}}(k_0,x)
&\le
R_{\bar{P}}(k_0,x)+B_{k_0}(x).
\end{aligned}
\end{equation}
Therefore,
\begin{equation}
\begin{aligned}
& R_{\bar{Q}}(k_1,x)-R_{\bar{Q}}(k_0,x) \\
&\ge
R_{\bar{P}}(k_1,x)-R_{\bar{P}}(k_0,x) \\
&\quad
-B_{k_1}(x)-B_{k_0}(x).
\end{aligned}
\end{equation}
Thus, if
\begin{equation}
R_{\bar{P}}(k_1,x)-R_{\bar{P}}(k_0,x)
>
B_{k_1}(x)+B_{k_0}(x),
\end{equation}
then
\begin{equation}
R_{\bar{Q}}(k_1,x)>R_{\bar{Q}}(k_0,x).
\end{equation}
This is a sufficient condition, not a necessary one.
It formalizes a condition under which an unconstrained expected-score
advantage of an instructional key is preserved after grammar projection. If the model is insensitive to key wording, then
\begin{equation}
R_{\bar{P}}(k_1,x)-R_{\bar{P}}(k_0,x)
\approx 0,
\end{equation}
and the condition is unlikely to hold. If the CoT-style key introduces a large projection tax, the condition may also fail. This offers one possible explanation for why key-level effects can vary
across models.

\subsection{An Optional Activation-Based Interpretation}

The sufficient condition above can be connected to reasoning-field activation. Let $\mathcal{R}_x$ denote the set of useful reasoning continuations for input $x$, and define
\begin{equation}
A_k = \Pr_{y\sim \bar{P}_{k,x}}[y\in\mathcal{R}_x].
\end{equation}
Let
\begin{equation}
\begin{aligned}
\mu_k^+
&= \mathbb{E}[\tilde{M}_k(y)
   \mid y\in\mathcal{R}_x], \\
\mu_k^-
&= \mathbb{E}[\tilde{M}_k(y)
   \mid y\notin\mathcal{R}_x].
\end{aligned}
\end{equation}
Then
\begin{equation}
R_{\bar{P}}(k,x)
=
A_k\mu_k^+ + (1-A_k)\mu_k^-.
\end{equation}
If the CoT-style key increases $A_k$ and the conditional quality gap $\mu_k^+-\mu_k^-$ remains positive, then the semantic gain term $R_{\bar{P}}(k_1,x)-R_{\bar{P}}(k_0,x)$ is expected to increase. This interpretation is not used as a theorem; it only explains why key wording can change the semantic benefit term in Equation~\ref{eq:sufficient_condition_key}.

\section{Decoding-Regime Sensitivity}
\label{app:temperature_sensitivity}

As a decoding-regime sensitivity analysis, we additionally repeated the
full factorial evaluation three times with \texttt{temperature}=0.8,
while keeping the models, datasets, prompts, schemas, constrained-decoding
backend, and all other inference settings fixed.

Tables~\ref{tab:temperature_sensitivity_gsm8k}
and~\ref{tab:temperature_sensitivity_math500} report the mean factorial
effect across the three runs, the sample standard deviation across runs,
and unadjusted 95\% paired item-level percentile bootstrap confidence
intervals computed from run-averaged item-level contrasts.

Several large GSM8K effects remain stable across runs. For example,
Qwen2.5-7B retains a positive schema-key effect, while Llama-3.2-3B
retains a large negative schema-key effect and a positive interaction.
At the same time, selected effects are sensitive to the decoding regime,
most notably for DeepSeek-R1-Distill-Qwen-7B.

\clearpage
\begin{table*}[p]
\centering
\small
\caption{
Factorial effects on GSM8K under stochastic decoding with
\texttt{temperature}=0.8 across three runs.
Values report mean effect $\pm$ run-level standard deviation and
unadjusted 95\% paired item-level bootstrap confidence intervals
(in percentage points).
}
\label{tab:temperature_sensitivity_gsm8k}

\renewcommand{\arraystretch}{1.15}
\setlength{\tabcolsep}{6pt}

\begin{tabular}{lccc}
\toprule
Model & $\Delta_{\mathrm{key}}$ & $\Delta_{\mathrm{prompt}}$ & $\Delta_{\mathrm{int}}$ \\
\midrule
Qwen2.5-3B
& \makecell[c]{$+2.60 \pm 0.69$ \\ $[+0.96, +4.25]$}
& \makecell[c]{$-0.73 \pm 2.28$ \\ $[-2.35, +0.91]$}
& \makecell[c]{$-4.22 \pm 2.60$ \\ $[-6.75, -1.72]$} \\

Qwen2.5-7B
& \makecell[c]{$+7.88 \pm 0.20$ \\ $[+6.44, +9.33]$}
& \makecell[c]{$+5.23 \pm 1.40$ \\ $[+3.74, +6.75]$}
& \makecell[c]{$-4.40 \pm 1.91$ \\ $[-6.27, -2.58]$} \\

DeepSeek-R1-Distill-Qwen-7B
& \makecell[c]{$-13.44 \pm 4.71$ \\ $[-15.72, -11.17]$}
& \makecell[c]{$+6.77 \pm 2.65$ \\ $[+4.98, +8.54]$}
& \makecell[c]{$+16.96 \pm 3.60$ \\ $[+13.98, +19.91]$} \\

Qwen2.5-Coder-14B
& \makecell[c]{$+1.87 \pm 0.37$ \\ $[+0.30, +3.44]$}
& \makecell[c]{$+2.15 \pm 0.54$ \\ $[+0.94, +3.39]$}
& \makecell[c]{$-0.71 \pm 0.49$ \\ $[-2.70, +1.26]$} \\

Llama-3.2-1B
& \makecell[c]{$+3.13 \pm 0.74$ \\ $[+1.67, +4.65]$}
& \makecell[c]{$+4.12 \pm 0.46$ \\ $[+2.91, +5.33]$}
& \makecell[c]{$-1.47 \pm 0.49$ \\ $[-3.51, +0.56]$} \\

Llama-3.2-3B
& \makecell[c]{$-13.62 \pm 0.64$ \\ $[-15.77, -11.52]$}
& \makecell[c]{$+4.09 \pm 1.34$ \\ $[+2.35, +5.81]$}
& \makecell[c]{$+16.48 \pm 2.42$ \\ $[+13.52, +19.43]$} \\

Llama-3.1-8B
& \makecell[c]{$+1.16 \pm 1.16$ \\ $[-0.58, +2.91]$}
& \makecell[c]{$+4.78 \pm 1.06$ \\ $[+3.18, +6.37]$}
& \makecell[c]{$-1.06 \pm 3.25$ \\ $[-3.29, +1.11]$} \\
\bottomrule
\end{tabular}
\end{table*}

\begin{table*}[p]
\centering
\small
\caption{
Factorial effects on Math500 under stochastic decoding with
\texttt{temperature}=0.8 across three runs.
Values report mean effect $\pm$ run-level standard deviation and
unadjusted 95\% paired item-level bootstrap confidence intervals
(in percentage points).
}
\label{tab:temperature_sensitivity_math500}

\renewcommand{\arraystretch}{1.15}
\setlength{\tabcolsep}{6pt}

\begin{tabular}{lccc}
\toprule
Model & $\Delta_{\mathrm{key}}$ & $\Delta_{\mathrm{prompt}}$ & $\Delta_{\mathrm{int}}$ \\
\midrule
Qwen2.5-3B
& \makecell[c]{$+1.87 \pm 2.32$ \\ $[-0.27, +4.07]$}
& \makecell[c]{$+0.67 \pm 1.42$ \\ $[-1.27, +2.60]$}
& \makecell[c]{$-2.07 \pm 2.01$ \\ $[-5.07, +1.00]$} \\

Qwen2.5-7B
& \makecell[c]{$+1.67 \pm 1.29$ \\ $[-0.53, +3.87]$}
& \makecell[c]{$+1.00 \pm 1.11$ \\ $[-1.00, +3.00]$}
& \makecell[c]{$-0.93 \pm 1.10$ \\ $[-3.93, +2.13]$} \\

DeepSeek-R1-Distill-Qwen-7B
& \makecell[c]{$-17.00 \pm 1.74$ \\ $[-20.53, -13.60]$}
& \makecell[c]{$+2.13 \pm 1.14$ \\ $[-0.07, +4.40]$}
& \makecell[c]{$+16.40 \pm 2.20$ \\ $[+12.13, +20.73]$} \\

Qwen2.5-Coder-14B
& \makecell[c]{$-3.47 \pm 0.83$ \\ $[-5.53, -1.40]$}
& \makecell[c]{$+0.13 \pm 1.36$ \\ $[-1.67, +2.00]$}
& \makecell[c]{$-0.67 \pm 2.50$ \\ $[-3.80, +2.33]$} \\

Llama-3.2-1B
& \makecell[c]{$-1.93 \pm 1.55$ \\ $[-3.40, -0.47]$}
& \makecell[c]{$-0.27 \pm 0.95$ \\ $[-1.67, +1.13]$}
& \makecell[c]{$+2.60 \pm 1.04$ \\ $[+0.67, +4.60]$} \\

Llama-3.2-3B
& \makecell[c]{$-5.00 \pm 1.56$ \\ $[-7.20, -2.87]$}
& \makecell[c]{$+2.40 \pm 2.75$ \\ $[+0.47, +4.33]$}
& \makecell[c]{$+4.53 \pm 2.48$ \\ $[+1.60, +7.53]$} \\

Llama-3.1-8B
& \makecell[c]{$+0.60 \pm 2.80$ \\ $[-1.67, +2.93]$}
& \makecell[c]{$+0.53 \pm 1.70$ \\ $[-1.67, +2.73]$}
& \makecell[c]{$-1.67 \pm 2.70$ \\ $[-4.73, +1.33]$} \\
\bottomrule
\end{tabular}
\end{table*}
\clearpage

\end{document}